%% file: main.tex
\documentclass[letterpaper, 10 pt, conference]{ieeeconf}  %
\usepackage{url}
\usepackage{booktabs}

\IEEEoverridecommandlockouts                              %

\input{macros.tex}

\title{\LARGE \bf
Maximal Adaptation, Minimal Guidance:\\
Permissive Reactive Robot Task Planning with Humans in the Loop
}

\author{
Oz Gitelson$^{1}$,
Satya Prakash Nayak$^{2}$,
Ritam Raha$^{2}$,
Anne-Kathrin Schmuck$^{2}$\\
$^{1}$Yale University, USA.\\
{\tt\small oz.gitelson@yale.edu}\\
$^{2}$Max Planck Institute for Software Systems, Kaiserslautern, Germany.\\
{\tt\small {sanayak,rraha,akschmuck}@mpi-sws.org}%
}

\begin{document}

\maketitle
\thispagestyle{empty}
\pagestyle{empty}

\begin{abstract}
We present a novel framework for human-robot \emph{logical} interaction that enables robots to reliably satisfy (infinite horizon) temporal logic tasks while effectively collaborating with humans who pursue independent and unknown tasks. The framework combines two key capabilities: (i) \emph{maximal adaptation} enables the robot to adjust its strategy \emph{online} to exploit human behavior for cooperation whenever possible, and (ii) \emph{minimal tunable feedback} enables the robot to request cooperation by the human online only when necessary to guarantee progress. This balance minimizes human-robot interference, preserves human autonomy, and ensures persistent robot task satisfaction even under conflicting human goals. We validate the approach in a real-world block-manipulation task with a Franka Emika Panda robotic arm and in the Overcooked-AI benchmark, demonstrating that our method produces rich, \emph{emergent} cooperative behaviors beyond the reach of existing approaches, while maintaining strong formal guarantees.
\end{abstract}

\section{Introduction}
 \input{sections/intro.tex}

\input{figures/grid}
\section{Problem Setup}\label{sec:problemSetup}
\input{sections/problemSetup.tex}

\section{Adaptability and Feedback Mechanisms via Permissive Strategy Templates}\label{sec:templates}
\input{sections/templates.tex}

\section{Experiments}\label{sec:experiments}
\input{sections/experiments.tex}

\bibliographystyle{IEEEtran}
\bibliography{main}

\end{document}

%% file: macros.tex
\usepackage{amsfonts}

\usepackage{amsthm}
\usepackage{mathtools}
\usepackage{xspace}
\usepackage{cite}
\usepackage{tikz}
\usepackage[percent]{overpic}
\usepackage{hyperref}
\usepackage{color}

\usepackage{url}

\newtheorem{remark}{Remark}
\newtheorem{proposition}{Proposition}
\newtheorem{problem}{Problem}

\theoremstyle{definition}
\newtheorem{example}{Example}
\newtheorem{definition}{Definition}

\usepackage[capitalise]{cleveref}
\crefname{definition}{Def.}{Def.}
\crefname{equation}{}{}
\crefname{figure}{Fig.}{Fig.}
\crefname{problem}{Prob.}{Prob.}
\crefname{algorithm}{Algo.}{Algo.}
\crefname{theorem}{Thm.}{Thm.}
\crefname{lemma}{Lem.}{Lem.}
\crefname{appendix}{App.}{App.}
\crefname{line}{line}{lines}
\crefname{item}{item}{items}
\crefname{section}{Sec.}{Sec.}
\crefname{corollary}{Cor.}{Cor.}
\crefname{proposition}{Prop.}{Prop.}
\crefname{example}{Ex.}{Ex.}
\crefname{remark}{Rem.}{Rem.}

\newcommand{\gridlabel}[1]{\mathtt{#1}}
\newcommand{\gridAdj}{\mathtt{adj}}
\newcommand{\gridDiag}{\mathtt{diag}}
\newcommand{\gridHalf}{\mathtt{major}}

\newcommand{\cupdot}{\mathbin{\mathaccent\cdot\cup}}

\newcommand{\tup}[1]{\left\langle #1\right\rangle}

\newcommand{\domain}{\mathcal{D}}
\newcommand{\states}[1]{{S_{#1}}}
\newcommand{\initState}{{s_0}}
\newcommand{\actions}[1]{{A_{#1}}}
\newcommand{\labelling}{L}
\newcommand{\ap}{\mathit{AP}}
\newcommand{\strat}[1]{{\pi_{#1}}}
\newcommand{\run}{\rho}

\newcommand{\true}{\top}
\newcommand{\globally}{\xspace\square\xspace}
\newcommand{\finally}{\xspace\lozenge\xspace}
\newcommand{\until}{\ U\ }
\newcommand{\nextt}{\xspace\bigcirc\xspace}

\newcommand{\game}{\mathcal{G}}
\newcommand{\acc}{\Omega}
\newcommand{\paritygame}[1]{{\mathit{Parity}[#1]}}
\newcommand{\col}{c}

\newcommand{\template}{\Pi}
\newcommand{\unsafe}{U}
\newcommand{\colive}{C}
\newcommand{\livegroups}{\mathcal{H}}
\newcommand{\live}{H}

\newcommand{\parameter}{\alpha}

%% file: sections/intro.tex
Effective human-robot interaction (HRI) requires robots to operate alongside humans who pursue their own goals -- often without explicitly revealing them. Such instances appear increasingly in domains ranging from smart manufacturing and logistics to assistive robotics in healthcare and households. In such scenarios, the robot must not only plan its moves to satisfy its own task but also adapt online to human behavior that may be cooperative, indifferent, or even obstructing its task. At the same time, HRI becomes more effective and enjoyable for the human, if robots do not only \emph{react} to what humans do, but respect and even leverage human behavior rather than always overwriting or constraining it. %

\begin{figure}
    \centering
    \begin{overpic}[width=0.48\textwidth]{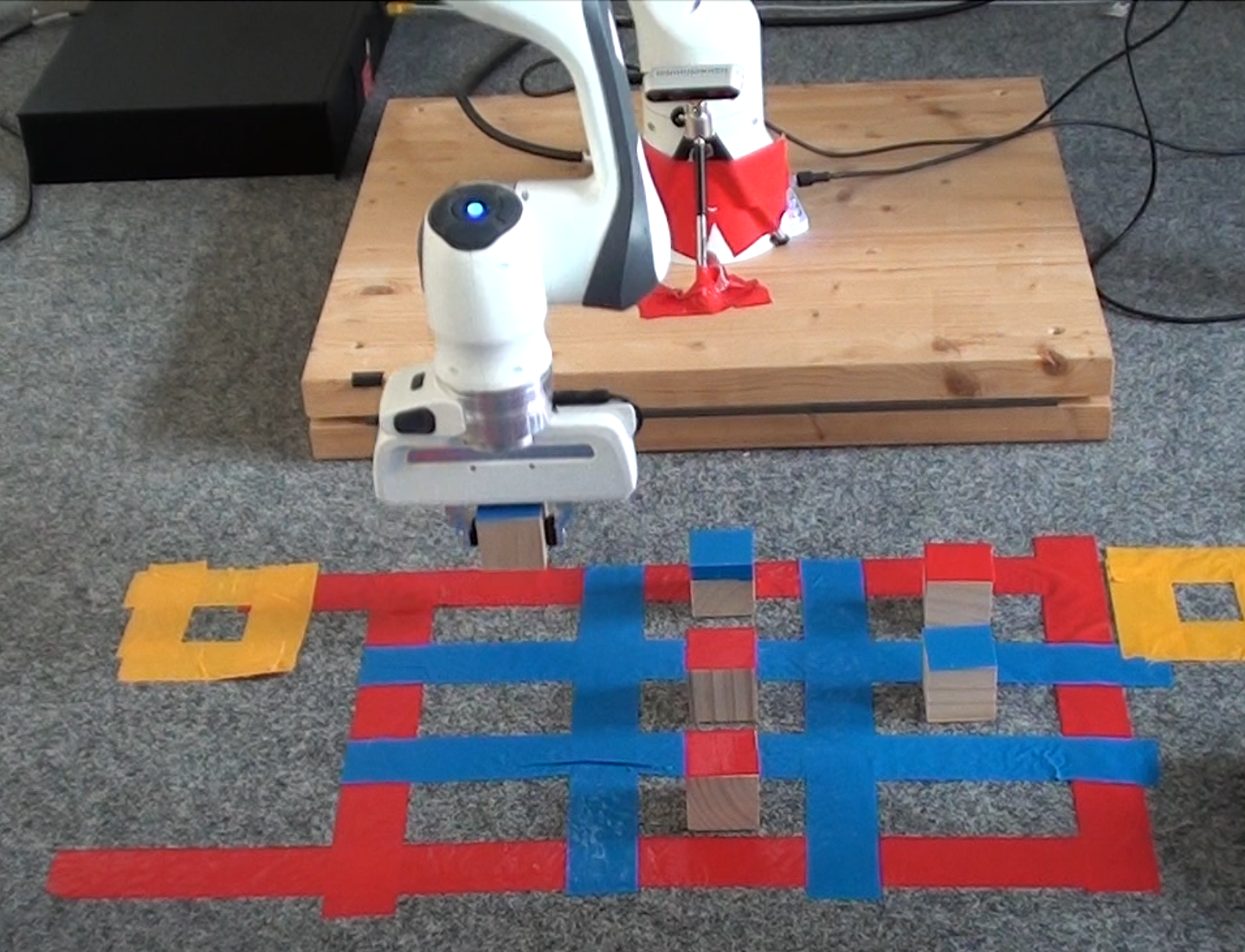}
        \put(0.5,49){\includegraphics[width=0.13\textwidth]{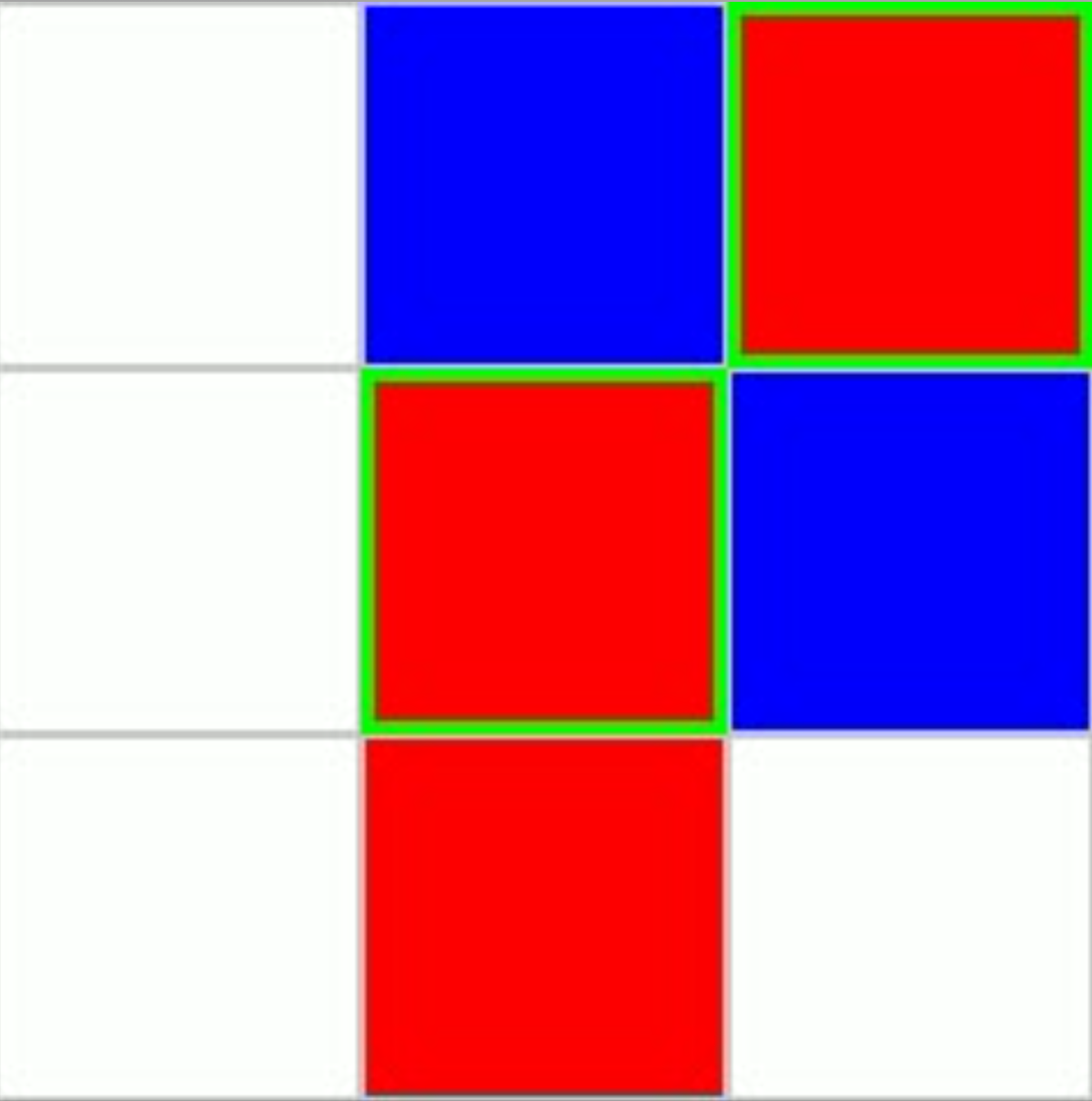}}
    \end{overpic}
    \caption{A simplified gridworld block-manipulation domain from our experimental setup, where a Franka Emika Panda robotic arm takes turns with a human to place blocks in a $3\times 3$ grid. The robot places blue blocks, while the human places red blocks. The top-left inset illustrates the robot's feedback to the human, suggesting the removal of the block in cell $(1,3)$ or $(2,2)$. A video of the experiment is available at \url{https://youtu.be/61thSZDj5Ks}.}
\label{fig:grid-screenshot}
\end{figure}

This paper addresses this challenge in the context of human-robot \emph{logical} interactions (HR$\ell$I)
where the robot is assigned a high-level temporal task, expressed in linear temporal logic (LTL), and the human simultaneously pursues an unknown strategic latent task. As an example, consider the simplified manipulation task depicted in \cref{fig:grid-screenshot}, where a Franka Emika Panda robotic arm takes turns with a human to place blocks in a \(3\times 3\) domain. The robots' task is to ensure that the majority of the cells is always eventually occupied with no adjacent cells filled, while the hidden human objective is to form a diagonal. If the domain of a logical task is restricted (as in this example), all possible strategic interactions of the human and the robot can be encoded in a two-player game graph, schematically illustrated in \cref{fig:grid}. Particular states in this graph fulfil the robots' (resp. humans') objective and are intended to be visited always again by the robot (resp. the human). 
However, as the robot and the human are both able to move blocks, they can obstruct the satisfaction of each others' goal. Importantly, this might already happen if both have the same goal, e.g., forming a diagonal. Here, the human can persist in forming the diagonal south-west-to-north-east, while the robot persist to form the south-east-to-north-west diagonal, resulting in a live lock. This problem is amplified if robot and human objectives differ, or are even incompatible.

To solve this problem, this paper introduces a novel HR$\ell$I framework which enables robots to persistently satisfy their logical tasks, while ensuring human autonomy whenever possible and requesting cooperation only when necessary.

\subsection{Related Work}
Due to the enormous relevance of reliable human-robot interaction (HRI) for trustworthy autonomy, there is an enormous body of work\footnote{Over two million results on Google Scholar searching for `human robot interaction' on 15.9.25.} in this research area.
We therefore only focus on HR$\ell$I scenarios described by a game between the robot and the human. This research line is rooted in the seminal papers \cite{KressGazitFainekosPappas2009,sadigh2018planning} which are motivated by the observation that LTL is a powerful specification language to describe strategic objectives such as traffic rules for autonomous driving \cite{AlthoffBeltaReviewFMCEforAutonomousDriving}, or robot navigation \cite{kress2018synthesisForRobots_Review}.

A natural formulation of HR$\ell$I is as a two-player game between the robot and its environment.
There exists a rich body of work on graph games \cite{bloem2018graph,jurdzinski98,energyy,ehrenfeucht1979positional,HELOUET2022104806,chatterjee2012energy,banerjee2023fast,anand2025fair} which can be leveraged to compute reactive strategies for the robot to fulfill diverse complex tasks.
Unfortunately, most solutions typically either over-constrain the robot or the human. In the first case, a robot strategy is computed which ensures the robots' objective against all human strategies. This, however, makes solutions overly restrictive and typically (as in the manipulation domain example above) fails to produce a robot strategy altogether. In the second case, full cooperation between the robot and the human are assumed, and a strategy is computed for both. Thereby, the human is forced into a very rigid, fully constrained behavior -- ensuring reliability but at the cost of suppressing human autonomy. 

A middle-ground is provided by various approaches which allow for increased human autonomy. When only \emph{logical safety} is concerned -- i.e., ensuring that no \emph{bad} strategic interaction happens between the human and the robot -- reactive shielding mechanisms \cite{SharmaSBPH0AF22,robinson2021smooth,li2021reactive,gundana2022event} which intervene with human behavior only to avoid such bad interactions, can be deployed. If, however, \emph{logical liveness} objectives are present -- i.e., requiring that something \emph{good} (e.g., forming a diagonal in the above example) eventually happens -- safety shielding is not sufficient to guarantee the satisfaction of the specification (as illustrated by the live-lock of human and robot trying to form a diagonal discussed before).

To mitigate these issues, many approaches explicitly model human behavior -- either by predicting from trajectories~\cite{junges2018model} or by representing it as a Markov Decision Process~\cite{feng2016synthesis} -- and integrate these models into the synthesis framework. Alternatively, HR$\ell$I can be directly modelled as a stochastic two-player game, as e.g.\ in \cite{10611623}. While this introduces local viability of human strategies via stochasticity, it does not capture the need for strategic human autonomy. %

To further improve human autonomy, admissibility-based methods, such as \cite{muvvala2024beyond, muvvala2024admissibility} can be used, which enable robots to adopt behaviors that remain robust against a broad range of human actions while still ensuring task satisfaction. Orthogonal to this approach, Schuppe et al.~\cite{Schuppe0LT23} focus on interactive advice, where the robot provides assume–guarantee style guidance to the human to support the satisfaction of a \emph{shared} objective with minimal cooperation by the human. However, these approaches require the robot to commit to a \emph{fixed pre-computed} strategy and rely either on stringent assumptions on human behavior, or on static, predefined forms of advice. %

A different, but related approach to HR$\ell$I only considers a (non-reactive) logical \emph{planning} objective for the robot and ensures safety of humans in its workspace by implementing the resulting plan of the robot via control-barrier functions (CBF), which act as a safety filter on the resulting underlying continuous robot dynamics \cite{10675313}. This approach was recently incorporated into cooperative HR$\ell$I frameworks, such as \cite{9561958,10610123,8619113}, where the (reactive) logical objectives of both the robot and the human are known, allowing an offline centralized game solution. Online human robot adaptations are only considered in the lower physical layer via CBFs but without any strategic autonomy. Similarly, recent dynamic game approaches to HRI, e.g.\ \cite{sadigh2016planning,peters-2023-contingency} are focusing on the immediate physical, rather than the long-term strategic adaptation and interaction of humans and robots.

In contrast to these approaches, we present an autonomy-driven approach to HR$\ell$I that focuses on high-level strategic interactions. While prior work with this focus, such as~\cite{muvvala2024beyond, muvvala2024admissibility, 10937075, Schuppe0LT23,10230080} result in pre-computed forms of cooperation or feedback and only consider fixed horizon objectives (as discussed before), our framework exploits the synergy between \emph{online} adaptation and \emph{tunable} feedback to generate complex \emph{emergent} cooperation behavior for \emph{finite and infinite} horizon LTL tasks. 
Formally, our novel framework is enabled by \emph{permissive strategy templates} \cite{strategyTemplates} for $\omega$-regular games (derived from the LTL objective $\varphi$) which allow to concisely represent infinitely many strategies.
Strategy templates have been applied to various contexts~\cite{APA,negotiation,NayakS24,SchmuckHDN24,AnandNS24,nayak2023context,abs-2504-16528,abs-2505-14689}, but to the best of our knowledge, we are the first to apply them to HR$\ell$I. In particular, we leverage recent results from \cite{negotiation} which allow to capture all human and robot strategies that guarantee satisfaction of $\varphi$ under minimal cooperation. This forms the theoretical basis for our novel online adaptation under strong formal guarantees.

\subsection{Contribution}
Our main contribution is a general framework for HR$\ell$I, which is applicable to the full class of LTL tasks and does not assume any particular strategic cooperation of the human. Conceptually, this framework ensures that the robot treats its interactions with a human not solely as a source of uncertainty to be constrained, but increasingly as a resource to be utilized.
When pursuing a high-level LTL task $\varphi$, the robot strategy (i) adapts at runtime to the humans' (i.e., forming a south-west-to-north-east diagonal as pursued by the human instead of insisting on forming the other one) to maximize cooperation whenever possible, and (ii) provides strategic feedback to the human only when the robots' adaptation alone cannot ensure progress towards the satisfaction of $\varphi$ (e.g., if the human persistently removes the middle block obscuring to form a diagonal the robot will ask the human to stop taking this move). This minimizes human-robot interference, maximizes cooperation and preserves human autonomy whenever possible, while still guaranteeing the eventual satisfaction of robot tasks if the human eventually listens to the provided feedback. %

We validate our approach both in simulation and on robotic hardware. In addition to the robotic manipulation domain implementation in \cref{fig:grid-screenshot}, we evaluate our framework in the \emph{Overcooked-AI} simulation environment~\cite{overcooked}, a widely used benchmark for collaborative planning with multiple actors. In this domain, the human and the robot repeatedly perform cooking tasks with the objective of persistently producing soups. Each participant is assigned an independent LTL task that encodes a recipe specification. As in the gridworld domain, these tasks are private and not known to the other. The human behavior is simulated by a probabilistic strategy.

In this application scenario the advantage of using $\omega$-regular specifications over commonly used reachability tasks becomes apparent. As both agents should produce as many specified soups as possible, they can actually cooperate even if their specifications are conflicting -- simply by 'taking turns' in producing the 'right' soup. We show that this intuitive cooperative behavior autonomously emerges via the online adaptation and feedback mechanism provided by our framework. To the best of our knowledge, the complexity of the emergent HR$\ell$I behavior provided by our framework thereby vastly exceeds the capabilities of all existing approaches while providing formal guarantees. %

%% file: figures/grid.tex
\usetikzlibrary{shapes,fit,positioning,calc}

\tikzset{
  human/.style={fill=red!60, draw=black, thick},
  robot/.style={fill=blue!60, draw=black, thick},
  gamegrid/.style={draw,gray,thin},
  live/.style={dashed, green!60!black, line width=2pt},
  final/.style={dotted, line width=1.2pt}
}

\newcommand{\gridsize}{0.3}   %
\newcommand{\tokenside}{0.53}  %
\newcommand{\tokenrad}{0.30}  %

\newcommand{\sgrid}[3]{%
  \begin{tikzpicture}[scale=\gridsize,baseline=(gridcenter)]
    \coordinate (gridcenter) at (1.5,-1.5);

    \foreach \i in {1,2} {
      \draw[gamegrid] (\i,0) -- (\i,-3);   %
      \draw[gamegrid] (0,-\i) -- (3,-\i); %
    }

    \pgfmathsetmacro{\halfside}{\tokenside/2}
    \foreach \r/\c in {#1} {
      \path[human] (\c-0.5,-\r+0.5)
        ++(-\halfside,-\halfside) rectangle ++(\tokenside,\tokenside);
    }

    \foreach \r/\c in {#2} {
      \path[robot] (\c-0.5,-\r+0.5) circle (\tokenrad);
    }

    \pgfmathsetmacro{\xhalf}{\tokenside/2}
    \foreach \r/\c in {#3} {
    \draw[black,thick,line width=1pt]
        (\c-0.5,-\r+0.5) ++(-\xhalf,-\xhalf) -- ++(2*\xhalf,2*\xhalf)
        (\c-0.5,-\r+0.5) ++(-\xhalf,\xhalf)  -- ++(2*\xhalf,-2*\xhalf);
    }
  \end{tikzpicture}%
}

\newcommand{\robnode}[4][]{%
  \node[ellipse,draw,thick,minimum width=1.2cm,minimum height=1.2cm,
        inner sep=1pt,label=above left:$\gridlabel{#2}$,#1] (#2) {%
    \sgrid{#3}{#4}{}
  };
}

\newcommand{\humannode}[4][]{%
  \node[rectangle,draw,thick,minimum size=1.2cm,
        inner sep=1pt,label=above left:$\gridlabel{#2}$,#1] (#2) {%
    \sgrid{#3}{#4}{}
  };
}

\newcommand{\targetnode}[3][]{%
  \node[minimum size=1.5cm,
        inner sep=1pt,
        label={[yshift=-8pt]above:$\gridlabel{#2}$},#1] (#2) {%
    \sgrid{}{}{#3}
  };
}

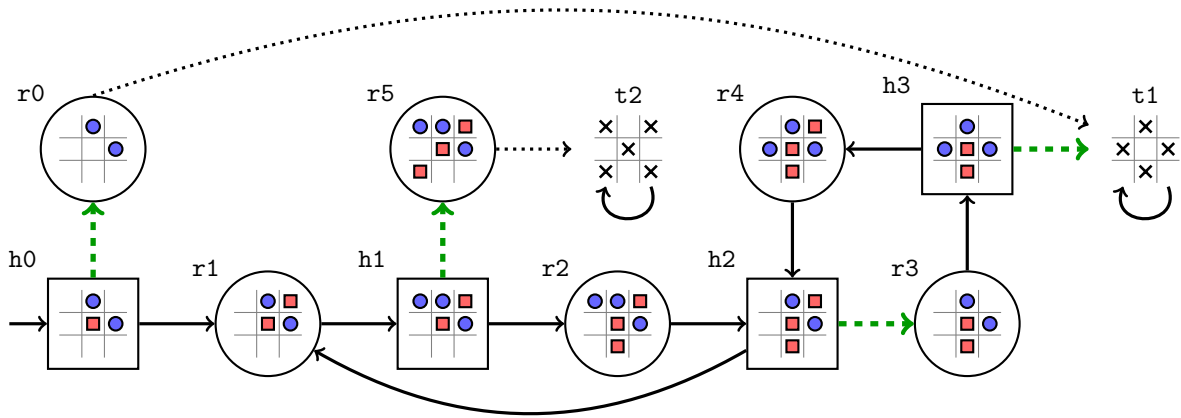
\begin{figure*}[t]
\centering
\begin{tikzpicture}[node distance=1cm,thick,line width=1.2pt] %

\humannode{h0}{2/2}{1/2,2/3}
\robnode[right=of h0]{r1}{2/2,1/3}{1/2,2/3}
\robnode[above=of h0]{r0}{}{1/2,2/3}

\humannode[right=of r1]{h1}{2/2,1/3}{1/1,1/2,2/3}
\robnode[right=of h1]{r2}{2/2,1/3,3/2}{1/1,1/2,2/3}
\robnode[above=of h1]{r5}{2/2,1/3,3/1}{1/1,1/2,2/3}

\humannode[right=of r2]{h2}{2/2,1/3,3/2}{1/2,2/3}
\robnode[right=of h2]{r3}{2/2,3/2}{1/2,2/3}
\humannode[above=of r3]{h3}{2/2,3/2}{1/2,2/3,2/1}

\robnode[left=of h3]{r4}{2/2,3/2,1/3}{1/2,2/3,2/1}

\targetnode[right=of h3]{t1}{1/2,2/1,2/3,3/2}
\targetnode[right=of r5]{t2}{3/1,2/2,1/3,1/1,3/3}

\draw[->] ($(h0.west)+(-0.5,0)$) -- (h0.west);
\draw[->] (h0) -- (r1);
\draw[->,live] (h0) -- (r0);
\draw[->] (r1) -- (h1);
\draw[->] (h1) -- (r2);
\draw[->,live] (h1) -- (r5);
\draw[->] (r2) -- (h2);
\draw[->,live] (h2) -- (r3);
\draw[->] (h2) to[bend left=30] (r1);
\draw[->] (r3) -- (h3);
\draw[->] (h3) -- (r4);
\draw[->,live] (h3) -- (t1);
\draw[->] (r4) -- (h2);

\draw[->, final] (r0.north) to[bend left=20] (t1);
\draw[->, final] (r5) -- (t2);
\draw[->] ($(t1)+(0.3,-0.5)$) to[loop below, in=250, out=290, distance=6mm] ($(t1)+(-0.3,-0.5)$);
\draw[->] ($(t2)+(0.3,-0.5)$) to[loop below, in=250, out=290, distance=6mm] ($(t2)+(-0.3,-0.5)$);

\end{tikzpicture}
\caption{An example of a partial reactive planning domain for turn-based human-robot interaction in a grid world. The robot controls the circle states, while the human controls the rectangle states. Each state contains a $3\times 3$ grid showing the current positions of the human-placed objects (red squares) and the robot-placed objects (blue circles). Directed edges represent possible actions leading to successor states. The robot's objective is to repeatedly reach states with majority-occupied cells where the placed objects are non-adjacent (i.e., no two occupied cells are neighbors), as illustrated in states~$\gridlabel{t1}$ and~$\gridlabel{t2}$. Green dashed edges denote \emph{live} actions, which are the suggested actions by the strategy templates, while dotted edges denote sequences of actions that lead to the target states.}
\label{fig:grid}
\end{figure*}

%% file: sections/problemSetup.tex
We focus on a turn-based human-robot interaction scenario, where the robot and the human alternately act in a shared environment. Given a high-level temporal task for the robot, our goal is to develop a framework that enables the robot to adapt its strategy online to the human’s observed behavior, and to provide feedback in a principled, tunable manner, so that the robot’s task is reliably satisfied even when the human pursues an independent objective whose chosen strategy may conflict with the robot’s progress.

\subsection{Reactive Planning Domain}\label{sec:problemSetup:domain}
We model the interaction between the robot and the human as a reactive planning domain $\domain = \tup{\states{},\initState,\actions{},\ap,\labelling}$, where:
\begin{itemize}
    \item $\states{} = \states{r}\cupdot \states{h}$ is a set of states, partitioned into robot states $\states{r}$ and human states $\states{h}$;
    \item $\initState \in \states{}$ is the initial state;
    \item $\actions{} = \actions{r}\cupdot \actions{h}$ is the set of actions (modeled as directed edges), partitioned into robot actions $\actions{r} \subseteq \states{r}\times \states{h}$ and human actions $\actions{h} \subseteq \states{h}\times \states{r}$;
    \item $\ap$ is the set of task-related propositions that can either hold or not hold in a given state.
    \item $\labelling\colon \states{} \to 2^{\ap}$ is a labeling function that labels each state with the set of propositions that hold in that state.
\end{itemize}
The planning domain can be specified using the planning domain description language (PDDL)~\cite{pddl}, a standard language in the field of AI planning. In a PDDL description, a state captures relevant objects and their locations, while actions are defined in terms of preconditions and effects.

A run $\run = s_0 s_1 s_2 \ldots$ of the planning domain is an infinite sequence of states such that $s_0 = \initState$ and for all $i \geq 0$, $(s_i, s_{i+1}) \in \actions{}$, i.e., there is an action that takes the system from state $s_i$ to state $s_{i+1}$.
The run $\run$ induces a \emph{trace} $\labelling(\run) = \labelling(s_0)\labelling(s_1)\labelling(s_2) \ldots$, an infinite word over $2^{\ap}$, which is a sequence of labels corresponding to the states in the run.
We assume that the robot and human take turns to act, i.e., if $s_i \in \states{r}$, then $s_{i+1} \in \states{h}$, and vice versa.

A robot \emph{strategy} $\strat{r}\colon \states{}^*\states{r} \to \actions{r}$ is a function that maps a sequence of states (representing the history of the interaction) ending in a robot state to the action that the robot should take.
A run $\run = s_0 s_1 s_2 \ldots$ is said to be $\strat{r}$-run if for all $i \geq 0$, whenever $s_i \in \states{r}$, then $s_{i+1} = \strat{r}(s_0 s_1 \ldots s_i)$.
A human strategy $\strat{h}$ and $\strat{h}$-run are defined similarly.

\begin{example}\label{ex:domain}
\cref{fig:grid} shows a partial view of a turn-based human–robot interaction modeled as a reactive planning domain. 
Each circle (e.g, $\gridlabel{r0}, \gridlabel{r1}, \ldots$) corresponds to a robot state in $\states{r}$ and each rectangle (e.g., $\gridlabel{h0}, \gridlabel{h1}, \ldots$) corresponds to a human state in $\states{h}$. 
Inside each node, the $3\times 3$ grid represents the environment: red squares denote human-placed objects, and blue circles denote robot-placed objects. 
Edges correspond to actions in $\actions{}$, alternating between human and robot moves according to the turn-based interaction. 
For example, the edge from state $\gridlabel{r2}$ to state $\gridlabel{h2}$ represents a robot action that removes a (blue circle) object from cell $(1,1)$ (i.e., 1st row, 1st column).
\cref{fig:grid-screenshot} shows an image of the same action being executed by a Franka robotic platform in a real-world setting.

A labeling function $\labelling$ can be defined to capture task-related propositions in $\ap$. 
For instance, let us consider the propositions $\ap = \{\gridAdj, \gridDiag, \gridHalf\}$, where $\gridAdj$ indicates that no two objects are adjacent (horizontally or vertically), $\gridDiag$ indicates that a diagonal is fully occupied, and $\gridHalf$ indicates that at least $4$ out of $9$ cells are occupied (majority-occupied).
In that case, $\gridAdj$ holds only in states $\gridlabel{r0}, \gridlabel{t1}, \gridlabel{t2}$; $\gridDiag$ holds in states $\gridlabel{r5}, \gridlabel{t2}$; and $\gridHalf$ holds in every state except $\gridlabel{h0}$ and $\gridlabel{r0}$.
From the initial state $\gridlabel{h0}$, a possible run is $\run = \gridlabel{h0}\,(\gridlabel{r1}\,\gridlabel{h1}\,\gridlabel{r2}\,\gridlabel{h2})^\omega$, which induces the trace $\labelling(\run) = \{\}\,\{\gridHalf\}^\omega$.
\end{example}

\subsection{Temporal Tasks as LTL formulas}\label{sec:problemSetup:ltl}
To express robot/human tasks, we use \emph{linear temporal logic} (LTL), a specification language that extends propositional logic with temporal operators~\cite{pnueli1977temporal}. 
Given a set $\ap$ of atomic propositions, LTL formulas are recursively defined as follows:
\[\varphi ::= \true \mid p \mid \neg \varphi \mid \varphi_1 \land \varphi_2 \mid \nextt \varphi \mid \varphi_1 \until \varphi_2\]
where $p \in \ap$ is an atomic proposition; $\neg$ and $\land$ are the boolean operators' negation and conjunction, respectively; and $\nextt$ and $\until$ are the temporal operators 'next' and 'until', respectively.
Other standard operators such as disjunction ($\lor$), implication ($\Rightarrow$), finally ($\finally$), and globally ($\globally$) can be derived from the above operators.
The semantics of LTL formulas are defined over infinite sequences of sets of atomic propositions in $(2^{\ap})^\omega$ and can be found in standard books~\cite[Chapter 5.1.2]{baier2008principles}.
We say a run $\run$ of a planning domain satisfies an LTL formula $\varphi$, denoted $\run \models \varphi$, if the trace induced by $\run$ satisfies $\varphi$.

\begin{example}\label{ex:taskLTL}
Consider again the reactive planning domain in \cref{ex:domain} with atomic propositions 
$\ap = \{\gridAdj, \gridDiag, \gridHalf\}$. 
Suppose the robot's task is to repeatedly reach states where no two objects are adjacent and at least four cells are occupied, which can be expressed by the LTL formula
$\varphi_r = \globally \finally (\gridAdj \land \gridHalf)$.
A human's task could be to repeatedly reach a state where a diagonal is fully occupied, expressed by the LTL formula
$\varphi_h = \globally \finally \gridDiag$.
A run that repeatedly visits states $\gridlabel{t1}$ satisfies both tasks, while a run that eventually only loops in state $\gridlabel{t2}$ only satisfies the robot's task.
\end{example}

\subsection{Problem Statement}
In this work, we are interested in the scenario of human-robot interaction where both the robot and the human have their own independent tasks, which are not known to each other.
In such settings, each agent may follow a strategy that, even if unintentionally, can block or halt progress toward the other's task.
Our goal is to develop a framework that enables the robot to adapt its strategy based on local observations and give feedback to the human when necessary, in order to persistently satisfy its own task over time.

\begin{problem}\label{prob:main}
    Given a human-robot interaction modeled as a reactive planning domain $\domain$ with an LTL task $\varphi$ for the robot in the presence of a human pursuing an unknown latent task, develop a framework that
    \begin{enumerate}[(a)]
        \item \textbf{captures} all cooperative behaviors of the human that enable the robot to satisfy its task $\varphi$;\label{item:prob:cooperative}
        \item \textbf{adapts} the robot's strategy based on local observations of the human's strategic behavior during interaction;\label{item:prob:adapt}
        \item \textbf{incorporates} a tunable mechanism for providing feedback to the human when needed, thereby facilitating the robot's progress toward fulfilling its task~$\varphi$.\label{item:prob:feedback}
    \end{enumerate}
\end{problem}

\begin{example}\label{ex:needForAdaptability}
Consider again the interaction in \cref{ex:domain}, where the robot's task is to repeatedly reach states with majority-occupied cells in which the placed objects are non-adjacent, as expressed by the LTL formula $\varphi_r = \globally \finally (\gridAdj \land \gridHalf)$ in \cref{ex:taskLTL}.
Suppose the human, pursuing its own latent task, repeatedly places objects along a diagonal in the grid. 
From the robot's perspective, this behavior can lead to a run satisfying $\varphi_r$, and thus this (unintentional) cooperation of the human should be exploited by the robot to fulfill $\varphi_r$.

However, in order to do so, the robot can not commit to a single strategy in advance to ensure that $\varphi_r$ is satisfied regardless of the human's behavior. To see this, let us assume that the robot choses to follow a fixed strategy which tries to satisfy $\varphi_r$ via a configuration as in $\gridlabel{t1}$. If the human still attempts to form a diagonal, their interaction will get stuck in a cycle where task $\varphi_r$ will never be satisfied.
Instead, the robot must \emph{adapt} its strategy by recognizing, from local observations, that the human is systematically filling the diagonal. It should then autonomously choose actions that lead to states where $\varphi_r$ is satisfied via a configuration as in $\gridlabel{t2}$. %
This, however, is only possible because $\varphi_r$ and the humans latent task can  indeed be satisfied simultaneously.

If, however, both objectives are conflicting, it does not suffice for the robot to adapt to the human behavior. As an example, consider a scenario where the human persistently takes actions $\gridlabel{h2} \to \gridlabel{r3}$ and $\gridlabel{h3} \to \gridlabel{r4}$. %
In such cases, the robot must recognize that the human's behavior can no longer be exploited for (unintended) cooperation. The robot then gives feedback to the human—e.g., requesting to remove the object at cell $(2,2)$ (by taking action $\gridlabel{h3} \to \gridlabel{t1}$).
\end{example}

%% file: sections/templates.tex
To reason about the strategic behavior of the robot for the LTL task, we reduce the planning domain along with the LTL task to a two-player game between the robot and human, as commonly done in the literature~\cite{baier2008principles}. We then leverage the recently developed notion of \emph{permissive strategy templates}~\cite{APA,strategyTemplates} in graph games to address \cref{prob:main}.

\subsection{$\omega$-Regular Games}
As a first step, we introduce the notion of two-player (turn-based) $\omega$-regular games, which will serve as the foundation for our framework.
\begin{definition}
    A two-player (turn-based) $\omega$-regular game is a pair $\game=\tup{\domain,\acc}$, where $\domain = \tup{\states{},\initState,\actions{},\ap,\labelling}$ is a reactive planning domain (as defined in \cref{sec:problemSetup:domain}), and $\acc\subseteq \states{}^\omega$ is an $\omega$-regular set of infinite sequences of states that defines the winning condition of the game.
\end{definition}

Such $\omega$-regular games can be canonically represented as \emph{parity games}~\cite{baier2008principles}.
A parity game is a special case of an $\omega$-regular game where the winning condition $\acc = \paritygame{\col}$ is defined by a coloring function $\col\colon \states{} \to \mathbb{N}$ that assigns a natural number (color) to each state, and a run $\run$ belongs to $\paritygame{\col}$ if the maximum color that appears infinitely often in $\run$ is even.
Using standard techniques~\cite{baier2008principles}, we can reduce a planning domain along with an LTL task to a parity game as formalized below.
\begin{proposition}\label{prop:reductionToGame}
    Given a reactive planning domain $\domain$ and an LTL formula $\varphi$ over $\ap$, we can construct a parity game $\game=\tup{\domain',\acc}$ such that there is a bijective correspondence between the runs of $\domain$ and the runs of $\domain'$, and a run $\run$ of $\domain$ satisfies $\varphi$ if and only if the corresponding run $\run'$ of $\domain'$ belongs to $\acc$.
\end{proposition}

\subsection{Permissive Strategy Templates}
In two-player games, a \emph{strategy template} generalizes the notion of a strategy by succinctly representing an infinite family of strategies through local constraints on the agent's actions. Formally, a strategy template $\template$ for agent $i$ consists of the following types of constraints:
\begin{itemize}
    \item \emph{Unsafe actions} $\unsafe\subseteq\actions{i}$: actions that the agent is prohibited from taking;
    \item \emph{Co-live actions} $\colive\subseteq\actions{i}$: actions that may only be taken finitely many times along any run;
    \item \emph{Live-groups} $\livegroups\subseteq 2^\actions{i}$: sets of actions such that, if the source state of some $\live\in\livegroups$ is visited infinitely often, the agent must take at least one action from the set infinitely often.
\end{itemize}
A run $\run$ is said to comply with a strategy template $\template$, 
if it satisfies all the specified constraints.
A strategy $\strat{}$ is said to follow a strategy template $\template$, denoted $\strat{} \models \template$, if all $\strat{}$-runs comply with $\template$.
For a detailed formal definition and further intuition, see~\cite{strategyTemplates}.

Recent work~\cite{negotiation} shows that, given a parity game between a robot and a human, it is possible to synthesize a strategy template for the human that captures all cooperative behaviors, and a corresponding strategy template for the robot that encompasses all strategies guaranteeing the winning condition against any cooperative human behavior.
\begin{proposition}\label{prop:negotiation}
    Given a parity game $\game = (\domain,\acc)$, a pair of strategy templates $(\template_r, \template_h)$ for the robot and human, respectively, can be synthesized such that 
    (i) every run $\run\in\acc$ also complies with $\template_h$; and
    (ii) every strategy $\strat{r}\models\template_r$ ensures that all $\strat{r}$-runs complying with $\template_h$ belong to $\acc$.
\end{proposition}

\begin{example}\label{ex:template}
For the grid world in \cref{fig:grid} with the robot's task $\varphi_r = \globally \finally (\gridAdj \land \gridHalf)$ as in \cref{ex:taskLTL}, a parity game can be constructed as per \cref{prop:reductionToGame} which has the same structure as the planning domain in \cref{fig:grid} but with an appropriate coloring function $\col$ that assigns all states where $\gridAdj \land \gridHalf$ holds the color $2$ (even) and all other states the color $1$ (odd).
This captures the robot's objective of repeatedly reaching states with majority-occupied cells where the placed objects are non-adjacent.
Using the synthesis procedure in \cref{prop:negotiation}, we can compute a pair of strategy templates $(\template_r, \template_h)$ for the robot and human, respectively, that capture cooperative behaviors.
For instance, the human's strategy template $\template_h$ includes live-groups $\{\gridlabel{h3} \to \gridlabel{t1}\}$ (in addition to other live groups as shown by green dashed edges in \cref{fig:grid}), which ensure that the human does not consistently obstruct the robot's ability to make progress as discussed in \cref{ex:needForAdaptability}.
Similarly, the robot's strategy template $\template_r$ includes live-groups that ensure the robot can always reach states satisfying $\gridAdj \land \gridHalf$ as long as the human follows $\template_h$.
\end{example}

\subsection{Adaptation and Feedback Mechanism}
While the results in \cref{prop:negotiation} provide a foundation for capturing cooperative behaviors, addressing \eqref{item:prob:cooperative} of \cref{prob:main}, they do not directly address the adaptation and feedback aspects outlined in \eqref{item:prob:adapt} and \eqref{item:prob:feedback} of \cref{prob:main}.
To this end, we propose a framework that leverages the permissive nature of strategy templates $(\template_r, \template_h)$ to enable the robot to adapt its strategy based on its strategy template $\template_r$ and provide feedback to the human based on the human's strategy template $\template_h$.

\smallskip
\noindent\textbf{Adaptation:} 
Since the template $\template_r$ provides a set of possible actions at each state, the robot does not need to commit to a single strategy in advance. Instead, at runtime, the robot randomly selects an action from the set of enabled actions in $\template_r$ at its current state. This approach allows the robot to adapt its choice of actions whenever it revisits a state.
In particular, if an action taken by the robot does not lead to a desirable outcome (e.g., the human appears to be uncooperative), the run will eventually return to the same state, and by randomness, the robot can try different actions from the enabled set in $\template_r$.
This adaptation mechanism effectively addresses \eqref{item:prob:adapt} of \cref{prob:main}.

\smallskip
\noindent\textbf{Feedback Mechanism:} 
To facilitate a tunable feedback mechanism as outlined in \eqref{item:prob:feedback} of \cref{prob:main}, we introduce a feedback threshold $\parameter \in [0,1]$ that determines how often the robot provides feedback to the human.
Note that, since unsafe actions in $\template_h$ are actions that the human must avoid satisfying the robot's task, the robot always communicates about unsafe actions whenever such an action is available at the current state.
For other types of constraints in $\template_h$, i.e., co-live actions and live-groups, the robot observes the human's actions and monitors how often the human violates these constraints (i.e., takes co-live actions or avoids actions from a live-group).
Whenever the frequency of such violations exceeds the feedback threshold $\parameter$, the robot provides feedback to the human from the next time step onward until the frequency of violations drops below $\parameter$.
Due to \cref{prop:negotiation}, as long as the human follows $\template_h$, the robot's strategy will ensure that the task is satisfied.
This feedback mechanism enables the robot to provide feedback in a tunable manner, only when human actions deviate significantly from the cooperative behaviors captured by $\template_h$, thus effectively addressing \eqref{item:prob:feedback} of \cref{prob:main}.

\begin{example}\label{ex:adaptationFeedback}
Continuing from \cref{ex:template}, the robot can adapt its strategy at runtime by randomly selecting actions from the enabled set in its strategy template $\template_r$.
For instance, suppose the human consistently places objects along the diagonal, as discussed in \cref{ex:needForAdaptability}.
Consider the scenario in \cref{fig:grid} where the robot is at state $\gridlabel{h0}$, and currently it has made progress toward reaching a state with configuration as in $\gridlabel{t1}$.
If the human continues to place objects along the diagonal, by taking actions $\gridlabel{h0} \to \gridlabel{r1}$ (which violates the live-group $\{\gridlabel{h0} \to \gridlabel{r0}\}$ in $\template_h$), the robot will adapt its strategy that now may lead to states like $\gridlabel{t2}$, which can accommodate the human's diagonal placements while still satisfying $\varphi_r$.
Now consider a scenario where the human, pursuing its own latent task, consistently places objects in a manner that obstructs the robot's task and loops between states $\gridlabel{h2}, \gridlabel{r3}, \gridlabel{h3}, \gridlabel{r4}$ as discussed in \cref{ex:needForAdaptability}.
In this case, once the frequency of violations of the live-groups exceeds the feedback threshold $\parameter$, the robot will provide feedback to the human, requesting to take the live action $\gridlabel{h3} \to \gridlabel{t1}$, which will help the robot make progress toward satisfying its task $\varphi_r$.
\end{example}

%% file: sections/experiments.tex
We evaluate the proposed framework on two experimental domains that illustrate complementary aspects of our novel HR$\ell$I framework. %
To demonstrate the power of our approach on a physical robotic platform, the first domain is the simplified gridworld block-manipulation setting depicted in \cref{fig:grid-screenshot} and described through our running example \cref{ex:domain,ex:needForAdaptability,ex:taskLTL,ex:template,ex:adaptationFeedback}. This example provides an interpretable test bed for visualizing how the robot adapts its strategy online in response to human actions and issues feedback in a tunable manner once the human blocks progress beyond a specified threshold. 

The second domain is the Overcooked-AI environment~\cite{overcooked}, a standard benchmark for collaborative planning, which highlights the power of guidance and adaptability to enable the fully autonomous emergence of complex strategic human robot interactions.
In particular, the natural $\omega$-regular structure of the specifications allows us to investigate different levels of difficulty for such emergent interactions, ranging from complete alignment to partial or full misalignment between the human's and the robot's tasks.

Together, these experiments showcase both the transparency of our method in symbolic domains and its power to produce complex emergent strategic human robot interactions fully autonomously at runtime, which go far beyond the capabilities of existing approaches.

\begin{remark}~\label{rem:FMA}
    Note that, \cite{Schuppe0LT23} also considers a human-robot interaction 'Follow My Advice' (FMA) scenario where the robot provides advice to the human. However, their approach focuses on computing sufficient assumptions for the human to device a static feedback mechanism based on a pre-computed robot strategy to achieve a finite-horizon goal. Our framework, in contrast, emphasizes online adaptability of the robot's strategy to align with the human's behavior and employs a tunable feedback mechanism to handle persistent $\omega$-regular objectives that we experimentally validate below. A faithful comparison would require re-formulating our settings to finite-horizon tasks, but this would undermine the core focus of our framework on the interplay between adaptability and feedback for persistent objectives.
\end{remark}

\subsection{Gridworld Block-Manipulation}
In this experiment, we implemented the simplified gridworld block-manipulation domain on a Franka Emika Panda robotic arm running ROS jazzy to demonstrate the feasibility of our framework beyond the abstract model. The setup consists of a robot hand operating on a $3\times 3$ workspace with tangible blocks that can be placed or removed, corresponding directly to the states illustrated in \cref{ex:domain}. The human interacts with the workspace by placing red blocks, while the robot places blue blocks. The system monitors the evolving configuration and evaluates whether the robot's task specifications (e.g., maintaining non-adjacent placements) are currently satisfied.

The reactive planning domain underlying this demonstration comprised approximately 7000 states and 18 propositions, encoding all possible placements of human and robot objects together with legal turn-taking moves. Our implementation required about 6 seconds to construct the parity game and synthesize the strategy templates, which was performed offline before execution. During execution, the robot follows its adaptive strategy: it updates its actions based on local observations of human moves, and when the human’s behavior risks blocking task satisfaction (as in \cref{ex:needForAdaptability}), the robot generates feedback through a display.

We include an image of the setup in \cref{fig:grid-screenshot} to illustrate how the abstract domain is realized in practice. This experiment highlights how our framework scales from the formal model to a real-world setting, providing an interpretable test bed to assess both adaptability and human feedback.

\subsection{Overcooked-AI}
We further evaluate our framework in the \emph{Overcooked-AI} environment, a widely used benchmark for collaborative planning with multiple actors. In this domain, the human and the robot repeatedly perform cooking tasks with the objective of persistently producing soups. Each participant is assigned an independent LTL task that encodes a recipe specification. As in the gridworld domain, these tasks are private and not known to the other. 

We consider three classes of experimental scenarios, characterized by the relation between the human's and the robot's recipes. In the first class, the recipes are identical, such that both participants unknowingly pursue the same task. In the second class, the recipes are distinct but compatible, meaning that there exists at least one type of soup that simultaneously satisfies both specifications. In the third class, the recipes are incompatible, i.e., there is no soup that satisfies both specifications simultaneously. These classes capture increasing levels of misalignment between the tasks pursued by the human and the robot. \cref{tab:recipes} summarizes the recipe configurations considered in our experiments. 

\begin{table}[h]
    \centering
    \caption{Recipe configurations in the Overcooked-AI experiments.}
    \label{tab:recipes}
    \begin{tabular}{lcc}
        \toprule
        \textbf{Scenario} & \textbf{Robot recipe} & \textbf{Human recipe} \\
        \midrule
        Identical & $=3$ onions & $=3$ onions \\
        Incompatible & $=3$ onions & $=2$ onions \\
        Compatible & $\geq 2$ onions & $\leq 2$ onions \\
        \bottomrule
    \end{tabular}
\end{table}

The Overcooked-AI environment provides a natural instantiation of persistent tasks, as both the human and the robot must repeatedly
complete recipes over time. Each recipe specification corresponds to an $\omega$-regular objective: an infinite run satisfies the task if
the required recipe is produced infinitely often. This allows us to evaluate not only whether a single goal is reached, but also whether
the tasks of the human and the robot can be persistently satisfied. 

We model the Overcooked-AI environment as a reactive planning domain as per \cref{sec:problemSetup:domain}, where the states encode the positions of the human and robot, the locations of ingredients, and the status of soups being cooked. The actions correspond to movement and interaction commands available to each participant. The $\omega$-regular tasks for the human and the robot are specified as a parity condition over the states, which can be derived from the recipe specifications.
The domain consists of approximately $68000$ states with over $200$ propositions encoding the relevant features of the environment.
We implement the synthesis procedure in \cref{prop:negotiation} to compute a pair of strategy templates $(\template_r, \template_h)$ for the robot and human, respectively, that capture cooperative behaviors, which took around $3$ minutes to compute offline before execution.
The robot then executes the adaptation and feedback mechanism as described in \cref{sec:templates} while the human is simulated by a probabilistic strategy for its recipe.

\begin{figure}[!h]
    \centering
    \includegraphics[width=0.48\textwidth]{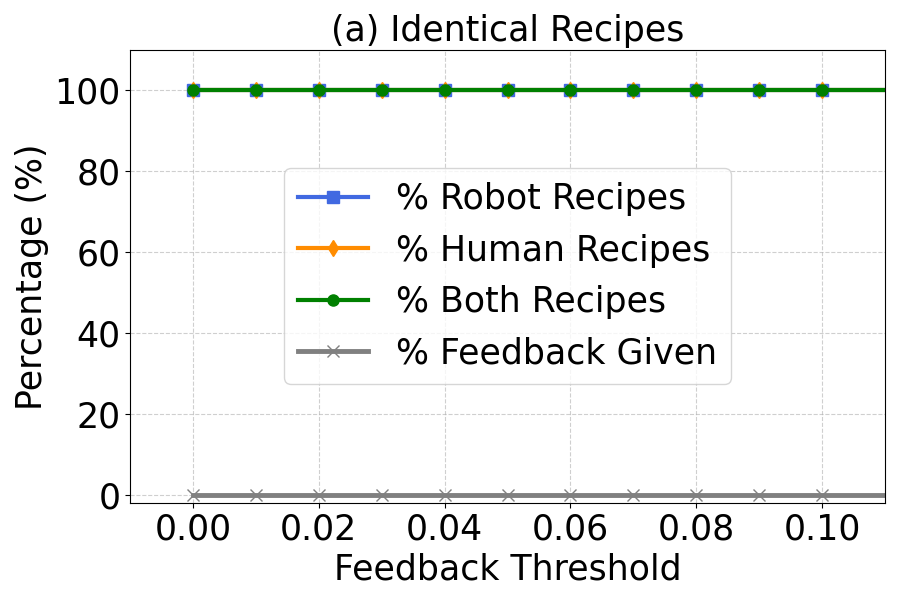}\\[1ex]
    \includegraphics[width=0.48\textwidth]{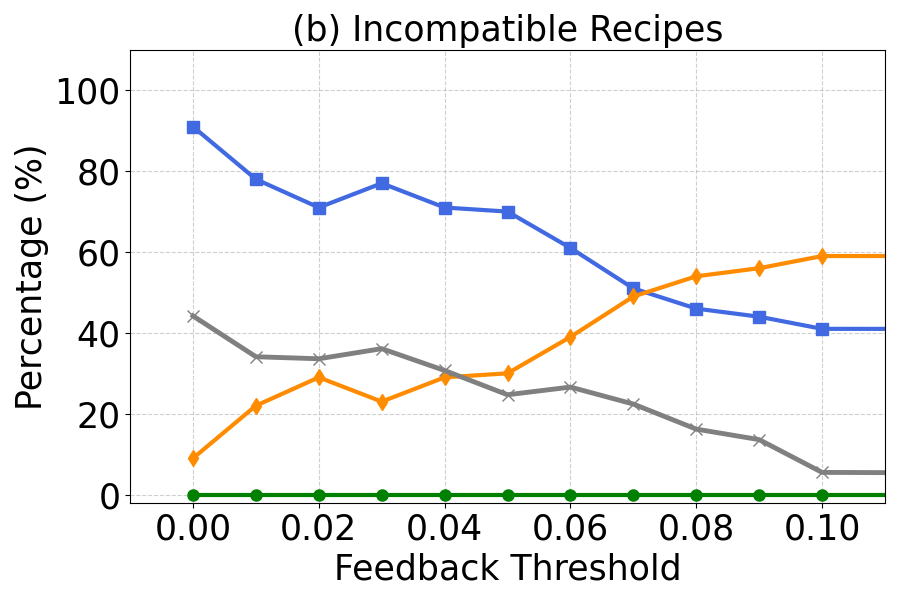}\\[1ex]
    \includegraphics[width=0.48\textwidth]{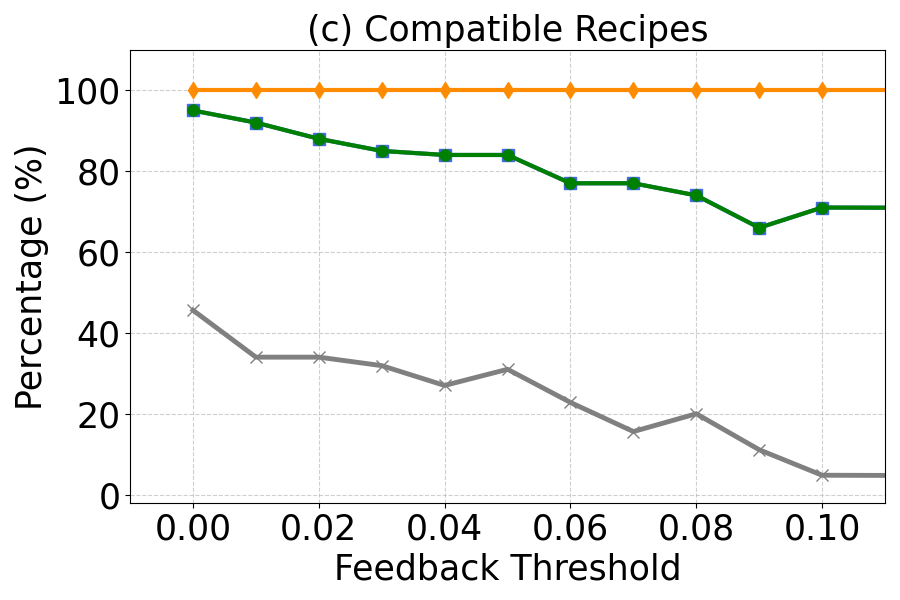}
    \caption{Satisfaction of human and robot recipe tasks over time in
    Overcooked-AI for (a) identical recipes, (b) incompatible
    recipes, and (c) compatible recipes. The plots show the
    proportion of runs satisfying each objective as well as frequency of feedback given.}
    \label{fig:buchi_results}
\end{figure}

We ran the three recipe scenarios (identical, compatible, incompatible) under different values of the feedback threshold $\alpha$ ranging from $0.00$ to $0.10$. For each scenario, the system was executed until $10$ soups were delivered. Each run lasted up to 500 moves with an execution time of $1$ minute and was repeated 10 times to account for the randomness in the human's and robot's action selection.
For each run, we recorded the following metrics: (i) the percentage of soups delivered that satisfy the robot's recipe, (ii) the percentage that satisfy the human's recipe, and (iii) the percentage that satisfy both recipes simultaneously. Additionally, we tracked the frequency of feedback issued by the robot throughout the run. \cref{fig:buchi_results} presents the temporal evolution of these metrics across all scenarios, illustrating how adaptability and feedback influence persistent task satisfaction.

\smallskip
\noindent\textbf{Identical recipes.} In the identical recipe setting, the human and the robot unknowingly pursue the same
recipe specification. As shown in \cref{fig:buchi_results} (a), both the human's and the robot's recipes are satisfied persistently across all runs. Crucially, no feedback is ever issued in this case. This illustrates the benefit of adaptability: even if the human and
the robot start with different strategies for producing the same recipe, the robot is able to adjust online so that their behavior
naturally aligns. In a system without adaptability, feedback would likely be needed to bring their strategies together,
whereas our framework allows cooperation to emerge autonomously at runtime. Note that this also showcases the advantage of our framework over the static feedback mechanism in \cite{Schuppe0LT23} (see \cref{rem:FMA}), which would issue feedback even when the human and robot have identical tasks, since it does not adapt the robot's strategy online.

\smallskip
\noindent\textbf{Incompatible recipes.} In the incompatible recipe setting, the human and the robot follow recipe specifications such
that no soup can satisfy both recipes simultaneously. Consequently, as shown in \cref{fig:buchi_results} (b), the number of soups delivered that satisfy both recipes (green) remains zero. %
However, as objectives are formulated as $\omega$-regular properties, it suffices to \emph{always eventually} deliver a soup which satisfies the human's or robots' recipe, respectively. Hence, both agents can still cooperate by 'taking turns' in producing a soup via the humans' (orange) and the robots' (blue) recipe. \cref{fig:buchi_results} (b) shows that this intuitive cooperative behavior indeed autonomously emerges and is influenced by the feedback threshold.
As the feedback
threshold $\alpha$ increases, the robot becomes more lenient to human non-cooperation, leading to a higher fraction of runs in which the
human's recipe is delivered while the robot's satisfaction decreases. This effect highlights the role of feedback tuning: with an
appropriately chosen threshold (in this case, $\alpha = 0.07$), both the human and the robot succeed in satisfying their respective recipes
approximately $50\%$ of the time persistently, demonstrating the importance of calibrated feedback sensitivity for the quality of the emerging cooperative behavior.

\smallskip
\noindent\textbf{Compatible recipes.} In the compatible recipe setting, the human and the robot pursue different recipe specifications, but there exists at least one soup that satisfies both. As shown in \cref{fig:buchi_results} (c), in our experiments the human persistently satisfies their recipe in all runs, while the robot adapts its strategy and issues feedback to ensure its own recipe is also satisfied. Even with higher thresholds, the robot manages to satisfy its recipe in more than $70\%$ of runs by adapting its strategy online towards the shared goal. When combined with more frequent feedback (i.e., lower thresholds), the robot can further increase the rate of joint satisfaction, achieving up to $95\%$ of runs where both recipes are satisfied. This demonstrates the synergy between adaptability and feedback tuning, which together enable persistent satisfaction of both recipes.